\documentclass[10pt,twocolumn,letterpaper]{article}

\usepackage{times}
\usepackage{epsfig}
\usepackage{graphicx}
\usepackage{amsmath}
\usepackage{amssymb}

\usepackage{xspace}
\makeatletter
\DeclareRobustCommand\onedot{\futurelet\@let@token\@onedot}
\def\@onedot{\ifx\@let@token.\else.\null\fi\xspace}

\usepackage{booktabs}

\usepackage{booktabs}       
\usepackage{amsmath}
\usepackage{amssymb}
\usepackage{times}
\usepackage{epsfig}
\usepackage{graphicx}
\usepackage{amsmath}
\usepackage{amssymb}
\usepackage{tabularx}
\usepackage{multirow}
\usepackage{subfigure}
\usepackage{comment}
\usepackage{float}
\usepackage{wrapfig}
\usepackage[ruled,vlined]{algorithm2e}
\usepackage{algorithmic}
\usepackage{bm}
\usepackage{color}

\usepackage[breaklinks=true,bookmarks=false]{hyperref}

\date{\vspace{-5ex}}

\begin{document}
\title{Making a Case for Learning Motion Representations with Phase}

\author{S. L. Pintea\\
Computer Vision Lab\\
Delft University of Technology\\
{\tt\small S.L.Pintea@tudelft.nl}
\and
J. C. van Gemert\\
Computer Vision Lab\\
Delft University of Technology\\
{\tt\small J.C.vanGemert@tudelft.nl}
}

\maketitle
\begin{abstract}
\vspace{-5px}
This work advocates Eulerian motion representation learning over the current standard Lagrangian optical flow  model. 
Eulerian motion is well captured by using phase, as obtained by decomposing the image through a complex-steerable pyramid.
We discuss the gain of Eulerian motion in a set of practical use cases: 
(i) action recognition,
(ii) motion prediction in static images,
(iii) motion transfer in static images and,
(iv) motion transfer in video.
For each task we motivate the phase-based direction and provide a possible approach.
\end{abstract}
\vspace{-11px}
\section{Introduction}
\begin{figure}
	\centering
	\includegraphics[width=0.30\textwidth]{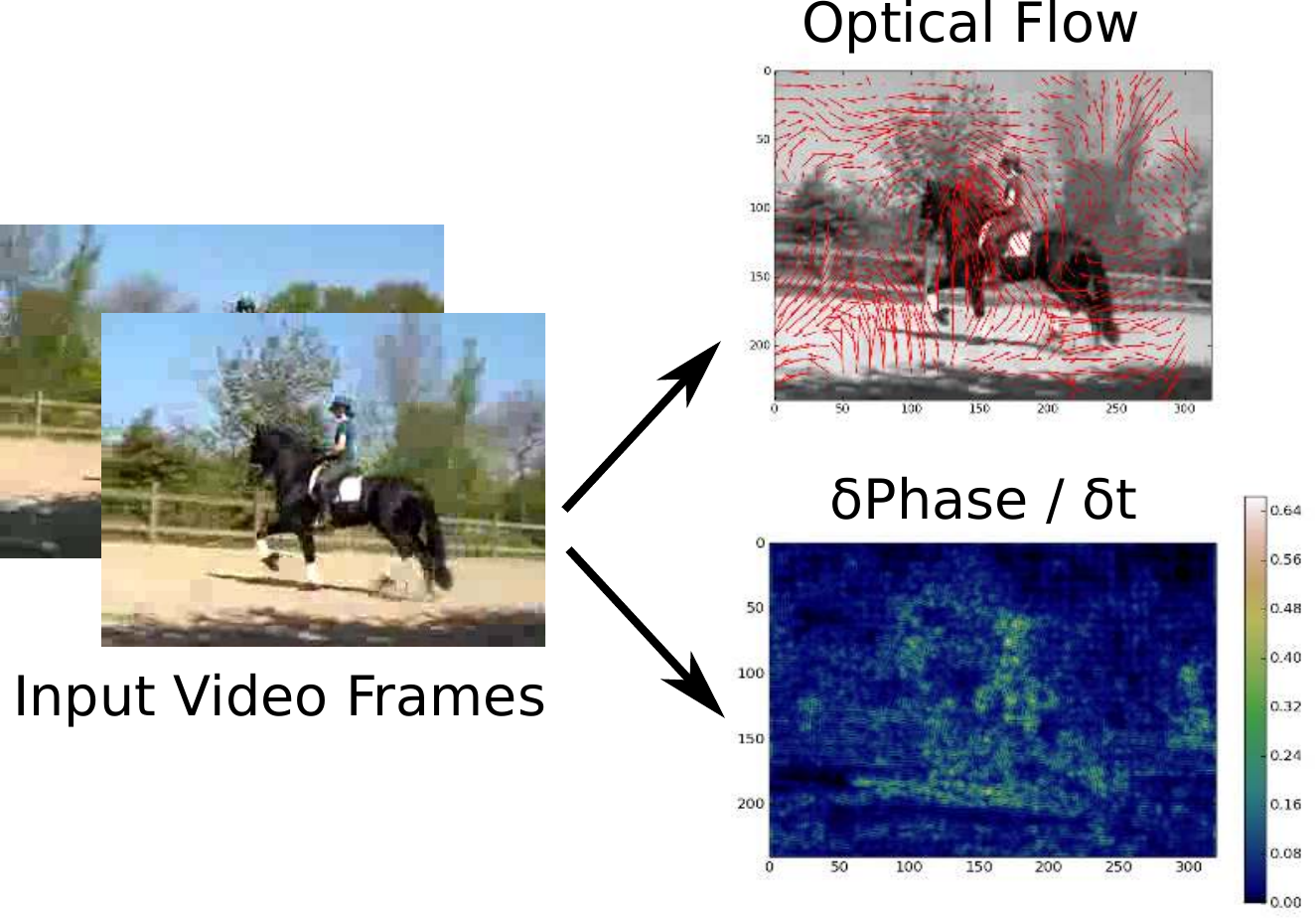}
	\vspace{-11px}
	\caption{\small While Lagrangian motion (optical flow) estimates the changes in position over time, it can miss correspondences or find mistaken correspondences. 
			However, in the Eulerian approach (phase variations over time) the number of motion measurements stays constant between frames, as for each input image we analyze the phase
			variations over time at each image location over multiple orientations and scales.} 
	\label{fig:eulerian}
	\vspace{-11px}
\end{figure}

We propose an Eulerian approach towards motion representation learning.  
The main difference between Lagrangian and Eulerian motion is that Lagrangian motion (optical flow) focuses on individual points
and analyzes their change in location over time. 
Therefore, Lagrangian motion performs tracking of points over time and for this it requires a unique matching method between point or patches.
On the other hand, Eulerian motion considers a set of locations in the image and analyzes the changes at these locations over time.
Thus, Eulerian motion does not estimate where a given point moves to, instead, it measures flux properties. 
Figure~\ref{fig:eulerian} depicts this difference between Eulerian and Lagrangian motion.
As a specific instance of the Eulerian model, we consider phase-based motion.
The phase variations over time of the coefficients of the complex-steerable pyramid are indicatives of motion 
\cite{fleet1990computation} and form the basis for learning motion representations. 

The gain of an Eulerian motion approach is that it avoids the need for hand-crafted optical flow constructions. Phase is an 
innate property of the image, it does not need to be estimated from explicit patch correspondences.
We propose a general-purpose phase-based motion description learning setup that can be used in any task relying on motion.
Here we explore four use cases:
(i) action recognition,
(ii) motion prediction in static images,
(iii) motion transfer in static images and,
(iv) motion transfer in video.
Note that phase-based motion representations are readily applicably to other motion-related task as well, including: human gait analysis, object tracking, action localization, etc.
\vspace{-11px}
\section{Related Work}
\vspace{-11px}
\subsection{Eulerian Motion} 
Eulerian motion modeling has shown remarkable results for motion magnification \cite{Wu12Eulerian} 
where a phase-based approach significantly improves the quality \cite{wadhwa2013phase} and broadens its application~\cite{chen2015modal,kooij2016depth}. A phase-based video interpolation is proposed in \cite{meyer2015phase} and a phase-based optical flow estimation is proposed in \cite{gautama2002phase}.
Inspired by the these work, we advocate the use of the Eulerian model as exemplified by phase for learning motion representations. 
\vspace{-11px}
\subsection{Action Recognition}
Optical flow-based motion features have been extensively employed for action recognition in works such as \cite{jain2013better,van2015apt,oneata2013action,wang2015action}.
These works, use hand crafted features extracted from the optical flow. 
Instead, we propose to input phase-based motion measurements to a CNN to reap the benefits of deep feature representation learning methods. 

A natural extension of going beyond a single frame  in a deep net is by using 3$D$ space-time convolutions~\cite{ji20133d,tran2014c3d}.
3$D$ convolutions learn appearance and motion jointly. 
While elegant, it makes it difficult to add the wealth of information that is available for appearance-only datasets through pre-training. 
In our method, we keep the benefit of pre-training by separating the appearance and the phase-based motion streams.

Using pre-trained networks is possible in the two-stream network approaches proposed in \cite{diba2016efficient,feichtenhofer2016convolutional,simonyan2014two}.
This combines a multi-frame optical flow network stream with an appearance stream and obtains competitive results in practice.
The appearance stream can employ a pretrained network.
Similarly, we also consider the combination of appearance and motion in a two-stream fashion,
but with innate phase information rather than using a hand-crafted optical flow.

The temporal frame ordering is exploited in \cite{fernando2015modeling}, where the parameters of a
ranking machine are used for video description. 
While in \cite{donahue2015long,li2016videolstm,srivastava2015unsupervised} recurrent neural networks are proposed for improving action recognition.
In this paper we also model the temporal aspect, although we add the benefit of a two-stream approach by separating appearance and phase variation over time. 

\vspace{-11px}
\subsection{Motion Prediction}
In \cite{pintea2014deja}, optical flow motion is learned from videos and predicted in static images in a structured regression formulation.
In \cite{walker2015dense} the authors propose predicting optical flow in a CNN from input static images. 
Where these works predict optical flow, we propose to predict the motion through phase changes, which does not depend on pixel tracking.

Predicting the future RGB frame from the current RGB frame is proposed in \cite{vondrick2015anticipating} in the context of action prediction. 
Similar to this work, we also start from an input appearance and obtain an output appearance image, however in our case the learning part learns 
the mapping from input phase information to future phase.
\vspace{-11px}
\subsection{Motion Transfer}
Animating a static image by transferring the motion from an input video is related to the notion of artistic style transfer \cite{gatys2016preserving,gatys2015neural,ruder2016artistic}. 
The style transfer aims at changing an input image or video such that the artistic style matches the one of a provided target image.
Here, instead, we consider the motion transfer --- given an input image, transfer the phase-based motion from the video to the image. 

Additionally, we also consider video-to-video transfer where the style of performing a certain action is transferred from a target video 
to the input video. 
In \cite{davis2015image} the authors allow the users to change the video by adding plausible object manipulations in the video.
Similar to this work, we also want to change the video motion after the recording is done, by adjusting the style of the action being performed. 
\vspace{-11px}
\section{Learning Motion with Phase}
\begin{figure}
	\centering
	\includegraphics[width=0.45\textwidth]{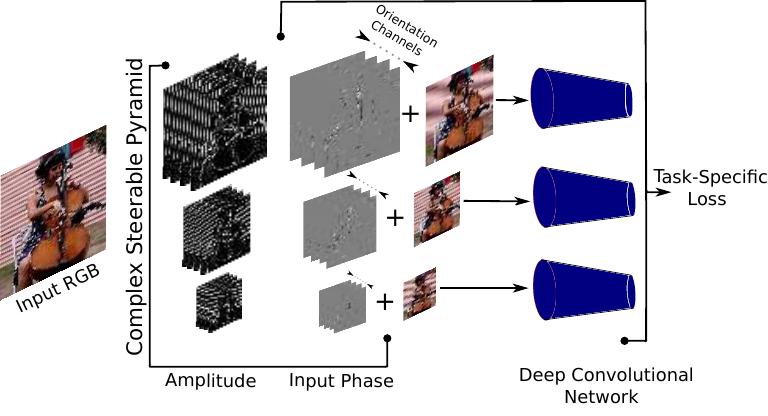}
	\vspace{-11px}
	\caption{\small Phase-based representation learning: from an input RGB image we extract phase information over multiple 
		orientations and scales by employing complex steerable filters. 
		For each scale, additional to the RGB input, we add the orientated phases as input to a network stream that optimizes a task-specific loss. 
	} 
	\label{fig:idea}
	\vspace{-11px}
\end{figure}
The local phase and amplitude of an image are measured by complex oriented filters of the form: 
$G_\sigma^\theta + i H_\sigma^\theta$, where $\theta$ is the filter orientation and $\sigma$ the filter scale \cite{freeman1991design},
\begin{equation}
	\small
	(G_\sigma^\theta + i H_\sigma^\theta) \otimes I(x,y) = A_\sigma^\theta (x, y) e^{i\phi_\sigma^\theta (x,y)}, 
\end{equation}
where $\phi_\sigma^\theta(x,y)$ is the local phase at scale $\sigma$ and orientation $\theta$, and $A_\sigma^\theta(x,y,t_0)$ the amplitude, 
$I(x,y)$ is the image brightness\slash input channel, and $\otimes$ the convolution operator, and $x, y$ are image coordinates. 
The filters have multiple scales and orientations, forming a complex steerable pyramid \cite{simoncelli1992shiftable} which captures various levels of image resolution.

There is a direct relation between motion and the change measured in phase over time.
The Fourier shift theorem makes the connection between the variation in phase of the subbands over time and the global image motion.
Rather than estimating global motion, using a steerable pyramid we can decompose the image into localized subbands and thus, recover the local motion in the phase 
variations over time. 
From the above decomposition only the phase, not the amplitude, corresponds to motion.
In \cite{fleet1990computation} the authors show that the temporal gradient of phase computed from a spatially bandpassed video over time, 
directly relates to the motion field.
Therefore, here, we focus on local phase at multiple scales and orientations to represent motion.

We propose using phase to learn motion representations for solving general motion-related tasks in a deep net.
We add phase as an additional motion input channel to a standard appearance (RGB) convolutional deep neural network.
Figure~\ref{fig:idea} shows our proposed general-purpose phase-based pipeline. 
The input video frame is decomposed using the complex steerable pyramid into amplitude and phase. 
Both phase and amplitude have multiple corresponding orientations and scales. Since the phase is an indicative of motion,
we ignore the amplitude and we use the input phase for the motion representation learning.
We treat the orientations as input channels while the scales represent different streams of the network, similar to \cite{denton2015deep} who use this setup for a different image pyramid. 

\vspace{-11px}
\section{Four Use Cases in Motion Learning}
We explore phase-based motion representation learning in four practical use cases. 
While a thorough in-depth experimental investigation is out of scope, we detail the setup of motion representation learning for each use case. 
\begin{figure*}[ht!]
	\centering
	\begin{tabular}{c@{\hskip 0.3in}c}
	\includegraphics[width=0.55\textwidth]{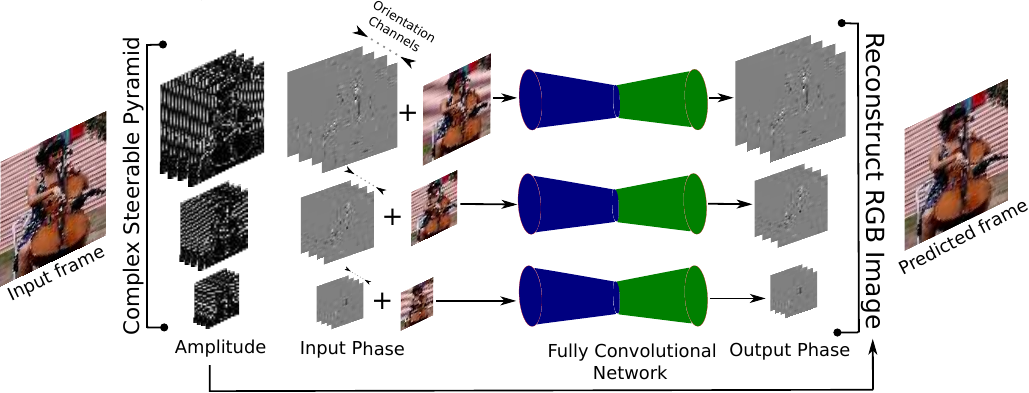} & 
	\includegraphics[width=0.25\textwidth]{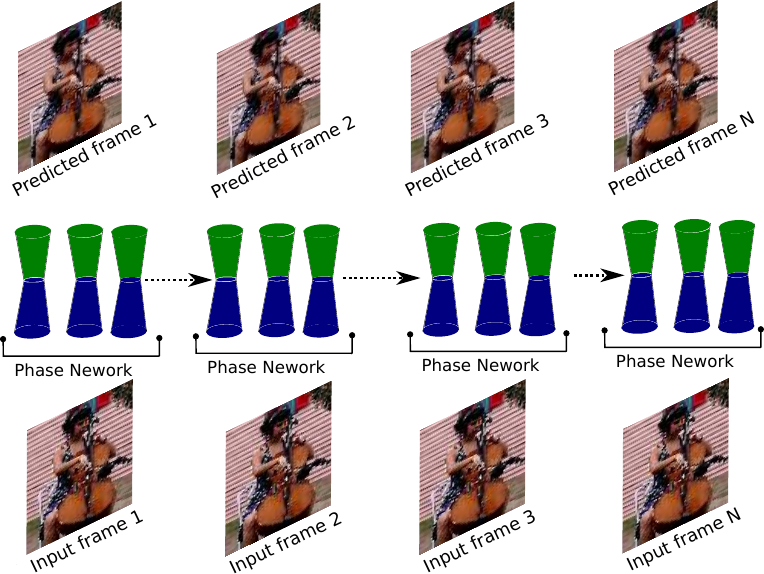}\\
	(a) Phase-motion prediction. & (b) Long term phase-based motion prediction.\\
	\end{tabular}
	\vspace{-11px}
    \caption{\small
		(a) Phase prediction in a Phase Network: from an input RGB image, we estimate the phase along multiple scales and orientations.
		For each scale we train a Fully Convolutional Network that predicts oriented phase at a future time-step. 
		From this we recover the predicted future RGB image.
		(b) Long-term motion prediction in static images: given the one step convolutional mapping from the input RGB image
		to the future RGB image, defined in the \lq Phase Network\rq, combine multiple of these networks in an Recurrent Neural Network to 
		obtain plausible long-term phase predictions.}
	\label{fig:phase_prediction}
	\vspace{-11px}
\end{figure*}
\vspace{-11px}
\subsection{Phase-based Action Recognition}
\begin{figure}
	\centering
	\includegraphics[width=0.45\textwidth]{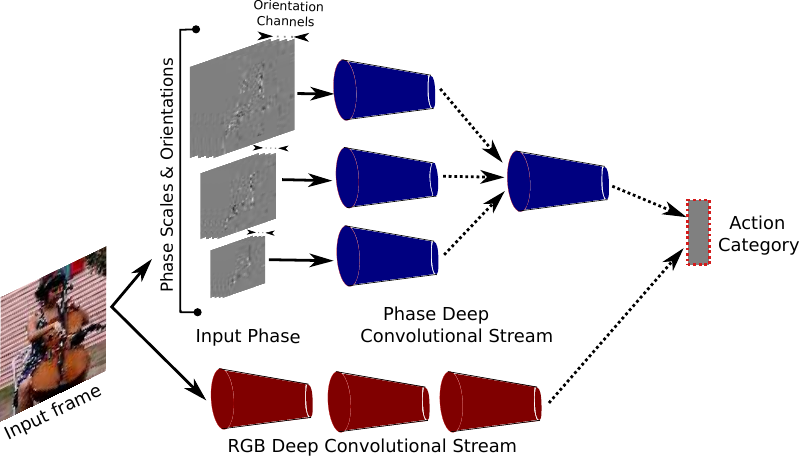}
	\vspace{-11px}
    \caption{\small Action recognition approach: two-stream CNN where the first stream receives input RGB frames,
		while the second stream receives input oriented phases of the video frame over multiple scales that are subsequently combined. 
	}
	\label{fig:action}
	\vspace{-11px}
\end{figure}

Separating appearance and motion in two-streams is effective for action recognition \cite{simonyan2014two}. 
For the appearance stream  we follow \cite{simonyan2014two} and use the input RGB frame, which offers the advantage of pre-training features on static images. 
However, where \cite{simonyan2014two} uses hand-crafted optical flow features, we propose to use Eulerian motion for the second stream with oriented phase over multiple scales, 
as depicted in figure ~\ref{fig:action}.

For evaluating action recognition, a comparison of our two-stream phase-based motion with the two-stream optical-flow approach 
of \cite{simonyan2014two} on the two datasets used in their paper --- HMDB51 and UCF101 is needed. 
We expect benefits from a phase-based motion representation because it does not depend on a specific hand-crafted optical flow implementation and does not rely on pixel tracking.
\vspace{-11px}
\subsection{Phase-based Motion Prediction in Static}
\label{ssec:phase_prediction}
The benefit of Eulerian motion for motion prediction is that the prediction locations are fixed over time. 
This contrasts sharply with Lagrangian motion, as pixels tracked by optical flow may be lost as they move in or out of the frame, or move to the same spatial location. 
Such lost pixels make it hard to recover long-term relations beyond just the next frame. 
The fixed prediction locations of a Eulerian motion representation do not suffer from this and offer long-term relation predictions of several frames. 

We propose to learn from a given input RGB the output future RGB, by recovering from the RGB the phase scales and orientations,
then predicting the multi-scale future phase-orientations and transforming them back into future RGB frames as in figure~\ref{fig:phase_prediction}.(a).  
For long-term motion prediction we propose an RNN (Recurrent Neural Network) version of this phase-based frame prediction, as depicted in figure~\ref{fig:phase_prediction}.(b). 
Thus, predicting motion $N$ timesteps away from the input.

For evaluating motion prediction we use the same datasets as in \cite{walker2015dense} --- HMDB51 and UCF101, 
where the authors aim at predicting optical-flow based motion in single images.
To evaluate the difference between the predicted motion and the actual video motion, we use pixel accuracy, as in our method 
we recover the appearance of the future frame. 
For comparison with \cite{walker2015dense} which reports EPE (End Point Errors), we use their chosen optical flow estimation 
algorithm to recover optical flow from our predicted RGB.
\vspace{-11px}
\subsection{Phase-based Motion Transfer in Images}
\label{ssec:transfer_im}
Similar to \cite{gatys2015neural, ruder2016artistic}, where the style of a given target painting is transferred to another image, we propose to transfer the short motion of a given video sequence to an input static image.
In \cite{gatys2015neural} a combination of two losses is optimized: content loss which ensures that the objects present in the newly generated image 
remain recognizable and correspond to the ones in the input image, and a style loss which imposes that the artistic style of the new image 
is similar to the one of the provided target painting. 
For motion transfer we have an additional requirement, namely that parts of the image that are similar --- e.g. horses, people, 
should move similar.
For this we use two pretrained network streams, an RGB stream and a phase stream and consider certain convolutional layers along these streams for 
estimation RGB\slash phase responses. 
Therefore, we first estimate an element-wise correlation between the responses at a given convolutional network layer 
of the input RGB values of the static image and the target video frame:
\begin{alignat}{1}
	\small	
	\mathcal{K}^l_j &= \frac{\sum_i^{N_l} C^l_{ij} D^l_{ij}}{\sqrt{\sum_i^{N_l} {C^l_{ij}}^2 }\sqrt{\sum_i^{N_l} {D^l_{ij}}^2 } },
\end{alignat}
where $N_l$ is the number of channels in the layer $l$, $C^l$ and $D^l$ the responses at layer $l$
for the input image and video frame, respectively. 
Following \cite{gatys2015neural}, we subsequently define our motion-style loss by weighting the feature maps in the Gram matrix 
computation by the appearance correlation.
The motion transfer is obtained by enforcing that the phase of objects over time in the input image, should be similar to the phase over time of the 
same objects present in the target video.
The motion-style loss optimization is performed per phase-scale.
\begin{alignat}{3}
	\small	
	G^l_\sigma[ij] &= \sum_k^{M_l} \mathcal{K}^l_k F^l_\sigma[ik] F^l_\sigma[kj],\text{ }i,j\in \{1,..N_l\},\\
	A^l_\sigma[ij] &= \sum_k^{M_l} \mathcal{K}^l_k P^l_\sigma[ik] P^l_\sigma[kj],\text{ }i,j\in \{1,..N_l\},\\
	\mathcal{L}_l &= \sum_\sigma \frac{1}{N_l^2 M_l^2}\sum_{i,j}^{N_l} \mathcal{K}^l_j (G^l_\sigma[ij] - A^l_\sigma[ij])^2,
\end{alignat}
where $M_l$ is the number of elements in one channel of layer $l$, $\sigma$ indicates the phase-scale, 
and $G^l$ is the weighted Gram matrix of the phase-image to be generated, 
while $A^l$ is the weighted Gram matrix of the current video frame and,
$F^l$ and $P^l$ are the responses of the phase-image to be generated and the phase-image of the input video frame, respectively.

Because we want the find similar looking objects by using the element-wise correlations, 
we expect that the higher convolutional levels of the network will perform better.
We additionally also add the content loss term of \cite{gatys2015neural} to avoid large distortions of the image appearance.
Due to the input being a static image, only short video motions can be transferred in this case.

For evaluating motion transfer, we perform a two-step evaluation.
In the first step, we select an existing video frame and transfer the video motion to the selected frame and compare the transferred motion 
with the actual video motion. 
For this we use videos from HMDB51 and UCF101. 
The second evaluation is transferring the motion to actual static images.
For this we select images from the static Willow dataset \cite{Delaitre10} and transfer the motion of corresponding 
videos from the HMDB51 and UCF101 datasets containing the same objects. 
For this we provide the static images animated with the transferred video motion.\\[5px] 
\vspace{-25px}
\subsection{Phase-based Motion Transfer in Videos}
\label{ssec:transfer_vid}
We use as a starting point the work of \cite{ruder2016artistic}, where artistic style is transferred to video. 
However in our case, the motion of one given video is transferred to another input video.
The gain in so doing, is that we can transfer the style of performing a certain action. 
For example an amateur performing the moonwalk can be lifted to the expert level by transferring the motion of Michael Jackson himself.  

The idea of transferring motion in videos is similar to the idea of transferring motion in static images, with the 
additional constraint that the motion must be temporally coherent.
For this, similar to \cite{ruder2016artistic}, we add a temporal loss term to the motion transfer loss discussed in section~\ref{ssec:transfer_im}. 

For performing motion transfer between videos, we use a set of target videos: 
the walk of Charlie Chaplin, the moonwalk of Michael Jackson, and the walk of a runway model.
We transfer these walking styles to a set of input videos of people walking, and provide the results as a qualitative form of evaluation.  

\vspace{-11px}
\subsection{Preliminary Proof of Concept}
Here\footnote{Demo: \href{http://silvialaurapintea.github.io/motion\_transfer/index.html}{http://silvialaurapintea.github.io/motion\_transfer/index.html
}.}, we show a very simple proof of concept for phase-based motion transfer. 
We animate a static image by transfering the motion of another semantic related video. 
Correctly aligning the moving entities between the video frames and the static image is essential for this task.  
For this proof of concept the alignment was not very good and no learning was used whatsoever. 
Misalignment errors show up as artifacts in the results and we expect that adding (deep) learning will improve results. 

\vspace{-11px}
\section{Conclusions}
We propose an Eulerian --phase-based-- approach to motion representation learning.
We argue for the intrinsic stability offered by the phase-based motion description. A phase-based approach does not require pixel tracking and directly encodes flux. Phase is an innate property of an image and does not rely on hand-crafted optical-flow algorithms.  We explore a set of motion learning tasks in an Eulerian setting:
(a) action recognition, 
(b) motion prediction in static images,
(c) motion transfer from a video to a static image and 
(d) motion transfer in videos.
For each one of these tasks we propose a phase-based approach and provide a small proof of concept. We do not offer in-depth experimental results but instead make a case for a brave new motion representation with phase.

\textbf{Acknowledgments.} This work is part of the research programme Technology
in Motion (TIM [628.004.001]), financed by the Netherlands Organisation for
Scientific Research (NWO).
\vspace{-11px}
{\small
\bibliographystyle{plain}
\bibliography{phasenet}
}
\end{document}